\documentclass[letterpaper,10 pt,journal,twoside]{ieeetrans}
\usepackage{ieee_robotics}
\usepackage{orcidlink}
\usepackage{siunitx}
\usepackage{mwe}
\usepackage{placeins}
\usepackage{progressbar}
\usepackage{ifthen}
\usepackage{xfrac}
\usepackage{bbm}
\usepackage{dblfloatfix}
\usepackage[normalem]{ulem}

\newcommand{\rev}[1]{#1}

\NewDocumentCommand{\mkProgBar}{ O{false} O{red} m }{
  \pgfmathsetmacro\percentage{#3}
  \pgfmathsetmacro\proportion{#3/100}
  \ifthenelse{\equal{#1}{true}} %
    {\underline{\textbf{\percentage\%}}} %
    {\percentage\%} %
  \progressbar[heightr=1.0,width=2.5em,filledcolor=#2!100]{\proportion}%
}

\newcommand{\GodDamnSensorName}{\texttt{\textbf{TacThru}}\xspace}

\newcommand{\GodDamnSystemName}{\texttt{\textbf{TacThru-UMI}}\xspace}

\acrodef{ddpm}[DDPM]{Denoising Diffusion Probabilistic Model}
\acrodef{mala}[MALA]{Metropolis-Adjusted Langevin Algorithm}
\acrodef{sdf}[SDF]{Signed Distance Function}
\acrodef{ibs}[IBS]{Intersection Bisector Surface}
\acrodef{fkine}[FK]{Forward Kinematics}
\acrodef{ik}[IK]{Inverse Kinematics}
\acrodef{ppo}[PPO]{Proximal Policy Optimization}
\acrodef{rl}[RL]{Reinforcement Learning}
\acrodef{gcrl}[GCRL]{Goal-Conditioned Reinforcement Learning}
\acrodef{mlp}[MLP]{Multi-Layer Perceptron}
\acrodef{cnn}[CNN]{Convolutional Neuron Network}
\acrodef{mdp}[MDP]{Markov Decision Process}
\acrodef{pd}[PD]{proportional-derivative}
\acrodef{ood}[OOD]{Out-of-Domain}
\acrodef{dfc}[DFC]{Differentiable Force Closure}
\acrodef{ar}[AR]{Augmented Reality}
\acrodef{llm}[LLM]{Large Language Model}
\acrodef{rwr}[RWR]{Reward-Weighted Regression}
\acrodef{vbts}[VBTS]{Vision-based Tactile Sensor}
\acrodef{sts}[STS]{See-Through-Skin}
\acrodef{umi}[UMI]{Universal Manipulation Interface}
\acrodef{tcp}[TCP]{Tool Center Point}
\acrodef{pdlc}[PDLC]{Polymer Dispersed Liquid Crystal}
\acrodef{pca}[PCA]{Principal Component Analysis}

\title{Simultaneous Tactile-Visual Perception for\\Learning Multimodal Robot Manipulation}

\author{
Yuyang Li\orcidlink{0000-0002-5794-7997}, Yinghan Chen\orcidlink{0009-0002-7110-8097}, Zihang Zhao\orcidlink{0000-0003-3215-7152}, Puhao Li\orcidlink{0009-0003-2696-9346}, Tengyu Liu\orcidlink{0000-0003-4006-1740}, Siyuan Huang\orcidlink{0000-0003-1524-7148}, Yixin Zhu\orcidlink{0000-0001-7024-1545} \\
\href{https://tacthru.yuyang.li}{\textcolor{magenta}{\texttt{\textbf{https://tacthru.yuyang.li}}}}

\thanks{This work is supported in part by the National Science and Technology Innovation 2030 Major Program (2025ZD0219400), the National Natural Science Foundation of China (62376009), the State Key Lab of General AI at Peking University, the PKU-BingJi Joint Laboratory for Artificial Intelligence, the Wuhan Major Scientific and Technological Special Program (2025060902020304), the Hubei Embodied Intelligence Foundation Model Research and Development Program, and the National Comprehensive Experimental Base for Governance of Intelligent Society, Wuhan East Lake High-Tech Development Zone.
We thank Lei Yan (LeapZenith), Shengyu Guo (PKU), Yu Liu (THU), and Leiyao Cui (PKU) for their assistance.}%
\thanks{Yuyang Li and Yinghan Chen contributed equally. Corresponding emails: \texttt{yixin.zhu@pku.edu.cn}, \texttt{huangsiyuan@ucla.edu}, and \texttt{liutengyu@bigai.ai}.}%
\thanks{Yuyang Li, Yinghan Chen, Zihang Zhao, and Yixin Zhu are with the Institute for Artificial Intelligence, Peking University.}%
\thanks{Yixin Zhu is also with the School of Psychological and Cognitive Sciences, Peking University.}%
\thanks{Yuyang Li, Yinghan Chen, Zihang Zhao, and Yixin Zhu are also with the Beijing Key Lab of Behavior and Mental Health, Peking University.}%
\thanks{Yuyang Li, Tengyu Liu, and Siyuan Huang are with the Beijing Institute for General Artificial Intelligence.}%
\thanks{All authors are with the State Key Lab for General Artificial Intelligence.}%
\thanks{Yixin Zhu is also with the Embodied Intelligence Lab, PKU-Wuhan Institute for Artificial Intelligence.}%
\thanks{Yinghan Chen is also with the Department of Computer Science and Technology, University of Cambridge.}%
}

\begin{document}

\let\oldtwocolumn\twocolumn
\renewcommand\twocolumn[1][]{%
    \oldtwocolumn[{#1}{
        \centering
        \includegraphics[width=\linewidth]{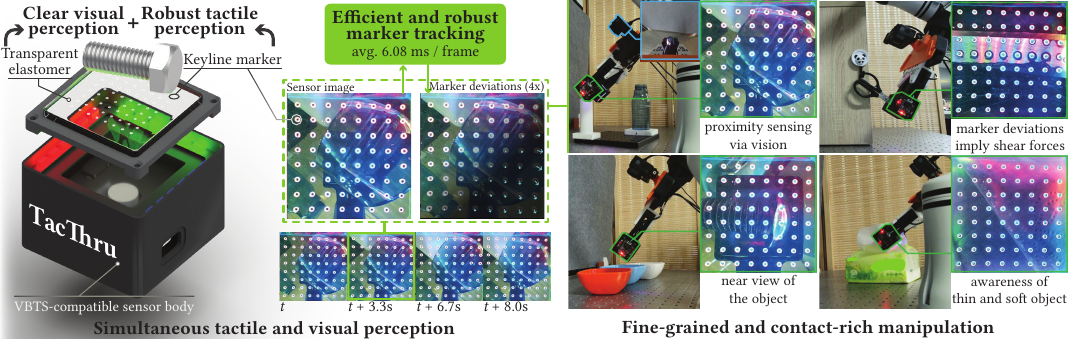}
        \captionof{figure}{\textbf{Learning multimodal robot manipulation with simultaneous tactile-visual perception.} \GodDamnSensorName{} enables clear visual perception and robust marker tracking via transparent elastomer and keyline markers (left), providing rich multimodal signals for learning manipulation policies. \GodDamnSystemName{} demonstrates efficacy across fine-grained and contact-rich tasks requiring precise multimodal coordination (right).}
        \label{fig:teaser}
        \vspace{6pt}
    }]
}

\markboth{}{Li and Chen \MakeLowercase{\etal}: TacThru}

\maketitle

\begin{abstract}
Robotic manipulation requires both rich multimodal perception and effective learning frameworks to handle complex real-world tasks. \Ac{sts} sensors, which combine tactile and visual perception, offer promising sensing capabilities, while modern imitation learning provides powerful tools for policy acquisition.
However, existing \ac{sts} designs lack simultaneous multimodal perception and suffer from unreliable tactile tracking. Furthermore, integrating these rich multimodal signals into learning-based manipulation pipelines remains an open challenge.
We introduce \GodDamnSensorName{}, an \ac{sts} sensor enabling \textit{simultaneous} visual perception and \textit{robust} tactile signal extraction, and \GodDamnSystemName{}, an imitation learning framework that leverages these multimodal signals for manipulation. Our sensor features a fully transparent elastomer, persistent illumination, novel keyline markers, and efficient tracking, while our learning system integrates these signals through a Transformer-based Diffusion Policy.
Experiments on five challenging real-world tasks show that \GodDamnSystemName{} achieves an average success rate of 85.5\%, significantly outperforming the baselines of tactile policy (66.3\%) and vision-only policy (55.4\%). The system excels in critical scenarios, including contact detection with thin and soft objects and precision manipulation requiring multimodal coordination.
This work demonstrates that combining simultaneous multimodal perception with modern learning frameworks enables more precise, adaptable robotic manipulation.
\end{abstract}


\section{Introduction}

\IEEEPARstart{M}{anipulation} requires a comprehensive environmental perception that spans the pre-contact to post-contact phases~\cite{billard2019trends,zhu2020dark}. Current sensing modalities each address different aspects but exhibit complementary limitations. Vision provides rich global context, but often downgrades during manipulation due to visual occlusions, particularly when precise control is required~\cite{zhao2025tac}. \Acfp{vbts}~\cite{yuan2017gelsight,lin20239dtact,ward2018tactip,li2024minitac} excel in providing high-resolution contact information~\cite{yuan2017gelsight,zhao2025tac,zhao2025embedding,zhao2025tacman}, but offer no information during the crucial pre-contact approach phase and provide only local, sometimes sparse signals~\cite{lancaster2022optical}. Dedicated proximity sensors partially bridge this pre-contact gap~\cite{jiang2012seashell,navarro2021proximity,fang2019toward}, but lack the spatial richness and precision needed for fine-grained manipulation.

\Acf{sts} sensors integrate tactile and visual sensing to address these limitations~\cite{yamaguchi2016combining,yamaguchi2017implementing,patel2018integrated,lancaster2019improved,hogan2022finger,lancaster2022optical,wang2022spectac,luu2023soft,luu2025vision,athar2023vistac,roberge2023stereotac,ablett2025multimodal,luo2024compdvision,fan2024vitactip,dong2025look}. Adapting \ac{vbts} designs, they replace opaque reflective coatings with semi-opaque or transparent alternatives, allowing the embedded camera to ``see through the skin'' for visual signals (\eg, object appearance, proximity) beside the tactile signals (\eg, contact, force). Recent work has validated their utility in object perception~\cite{athar2023vistac,fan2024vitactip,luo2024compdvision}, grasping~\cite{shimonomura2016robotic,athar2023vistac,lancaster2019improved,yamaguchi2017implementing}, in-hand translation~\cite{lancaster2022optical}, and articulated object manipulation~\cite{hogan2022finger,ablett2025multimodal}.

Despite this progress, three limitations prevent the broader adoption of \ac{sts} sensors. First, most current designs explicitly switch between tactile and visual modalities through illumination control~\cite{roberge2023stereotac,ablett2025multimodal} or movable components~\cite{dong2025look}, preventing \textit{simultaneous} multimodal perception and introducing additional control complexity. Second, tactile markers essential for shear force measurement are difficult to track against \textit{noisy}, \textit{unpredictable} external backgrounds, limiting applications in open environments~\cite{yamaguchi2016combining}. Finally, existing \ac{sts} applications rely primarily on \textit{hand-crafted} controllers, with integration into modern data-driven manipulation pipelines largely unexplored.

To address these issues, we introduce \GodDamnSensorName{}, an \ac{sts} sensor that enables simultaneous tactile-visual perception through (i) a fully transparent elastomer for clear visual perception, (ii) persistent illumination that eliminates mode switching, (iii) novel keyline markers robust to background variation, and (iv) an efficient tracking algorithm that enables 120Hz operations. This design enables truly \textit{simultaneous} tactile-visual perception while remaining compatible with standard \ac{vbts} fabrication~\cite{cui2025vi} with minimal adoption.

We further develop \GodDamnSystemName{}, integrating \GodDamnSensorName{} into an imitation learning framework based on \ac{umi}~\cite{chi2024universal}. A Transformer-based Diffusion Policy learns to properly attend to the \textit{simultaneously provided} multimodal signals for manipulation control. We evaluate the system on five diverse real-world tasks, including pick-and-place, sorting, and insertion (\cref{fig:teaser}), where \GodDamnSensorName{} provides persistent environment, object, and contact perception. Our policies reach an average success rate of 85.5\%, achieving $1.54\times$ and $1.29\times$ performance compared to our vision (55.4\%) and tactile (66.3\%) baselines, respectively.

Our contributions include: (i) \GodDamnSensorName{}, a novel \ac{sts} sensor that enables efficient, robust, simultaneous tactile-visual perception; (ii) \GodDamnSystemName{}, an imitation learning system with a design compatible with \ac{umi} for data collection, processing, and policy deployment; and (iii) a comprehensive experimental validation demonstrating how \GodDamnSensorName{}'s simultaneous multimodal perception enables superior fine-grained and contact-rich manipulation.

\section{Related Work}

\paragraph*{Tactile and \ac{sts} Sensors}
Tactile sensing enables robots to perceive fine-grained contacts during manipulation. Piezoresistive~\cite{yang2021non,yu2022all}, piezoelectric~\cite{tian2019rich,lv2022flexible}, and conductive-fluid~\cite{fishel2012sensing} sensors measure dynamic contact forces with compact forms, while \acf{vbts}~\cite{yuan2017gelsight,lin20239dtact,li2024minitac,suresh2024neuralfeels,lambeta2020digit,zhao2025embedding} provide high spatial resolution at relatively low cost. However, traditional tactile sensors are inherently limited to in-contact perception. \Ac{sts} sensors address this limitation by integrating tactile and visual perception~\cite{ablett2025multimodal,wang2022spectac,yamaguchi2016combining,hogan2022finger,athar2023vistac,luo2024compdvision,roberge2023stereotac,luu2023soft,luu2025vision}, typically replacing opaque coatings of \acp{vbts} with semi-opaque or transparent alternatives~\cite{ablett2025multimodal,hogan2022finger,roberge2023stereotac} for continuous perception before, during, and after contacts. Existing implementations commonly \textit{alternate} between tactile and visual modes using illumination control~\cite{wang2022spectac,athar2023vistac,hogan2022finger,roberge2023stereotac}, mechanically actuated elastomers~\cite{dong2025look}, or \acs{pdlc} films~\cite{luu2023soft,luu2025vision}. Other extensions incorporate UV-excited fluorescent markers~\cite{wang2022spectac,luo2024compdvision}, and stereo vision~\cite{roberge2023stereotac,shimonomura2016robotic,luo2024compdvision}.
Such mode switching often relies on task-specific heuristics, introducing control complexity and potential errors (\eg, millimeter-level error in sensor-object distance estimation~\cite{dong2025look,hogan2022finger}; second-level delays during actuation~\cite{dong2025look,ablett2025multimodal}). \GodDamnSensorName{} enables simultaneous tactile-visual perception with a fully transparent elastomer, persistent illumination, and robust keyline markers with efficient tracking.

\paragraph*{Multimodal Manipulation with \ac{sts} Sensors}
\ac{sts} sensors' visual-tactile sensing mitigates the inherent limitations of individual sensing modes: tactile feedback provides precise contact information but remains local and sparse~\cite{lancaster2022optical}; vision offers a wider spatial context but fails during occlusion or contact. This multimodal approach improves object pose estimation~\cite{shimonomura2016robotic,fan2024vitactip,athar2023vistac}, slip detection~\cite{yamaguchi2017implementing}, material recognition~\cite{fang2019toward}, in-hand manipulation~\cite{lancaster2022optical}, and visual servoing~\cite{hogan2022finger,athar2023vistac,ablett2025multimodal,dong2025look}, allowing unified sensor solutions for object localization, approach, and interaction. We further demonstrate \ac{sts} sensors' utilities in a series of diverse real-world tasks via \GodDamnSensorName{}.

\paragraph*{Learning with Multimodal Sensing}
Learning-based manipulation policies encode multimodal sensory streams into unified representations or tokens. Visual data processing employs \ac{cnn} or Transformer encoders~\cite{levine2016end,kalashnikov2018scalable,wu2020learning,shridhar2023perceiver,goyal2023rvt}, while tactile integration strategies vary by signal types: low-dimensional inputs (\eg, piezoelectric contacts) are processed via \acp{mlp}~\cite{liu2025vitamin,heng2025vitacformer,zhao2025touch}, while \acp{vbts} utilize specialized encoders for high-resolution images~\cite{radford2021learning,mees2022calvin,xu2025unit,zhao2024transferable} or extracted features (\eg, contact depth, marker displacement)~\cite{chen2024general,xu2023efficient}. Despite these advances, \ac{sts} sensors' integration into learning frameworks remains largely unexplored~\cite{ablett2025multimodal}. We integrate \GodDamnSensorName{} into an imitation learning framework,  \GodDamnSystemName{}, and demonstrate that simultaneous multimodal perception enables superior performance in fine-grained and contact-rich manipulation.

\section{\GodDamnSensorName{}}

\begin{figure}[!b]
    \centering
    \includegraphics[width=\linewidth]{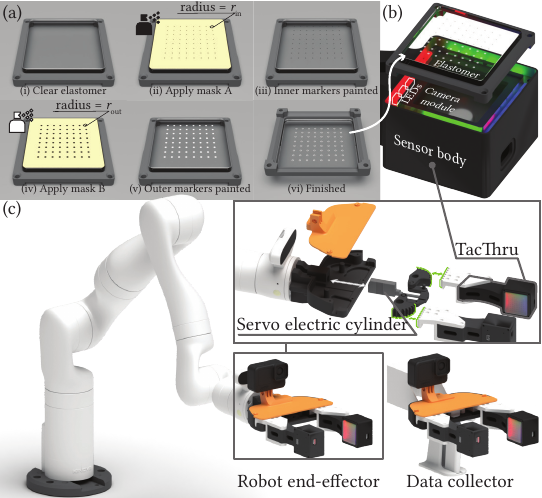}
    \caption{\textbf{\GodDamnSensorName{}'s fabrication and integration in \GodDamnSystemName{}.} (a) The keyline markers are fabricated by sequentially spraying inner (black) and outer (white) markers on a transparent elastomer. (b-c) \GodDamnSystemName{} includes a robot end-effector (left) and a data collector (middle) with identical body and \GodDamnSensorName{} fingers. The end-effector's fingers are actuated by a servo electric cylinder (right).}
    \label{fig:tacthru-umi-system}
\end{figure}

\GodDamnSensorName{} achieves simultaneous tactile-visual perception through three design principles: fully transparent elastomer for clear visual access, persistent illumination to eliminate modal switching, and robust keyline markers for reliable tactile tracking. \GodDamnSensorName{} maintains compatibility with the standard \ac{vbts} fabrication pipeline, differing primarily in elastomer material to allow easy adoption.

\subsection{Transparent Elastomer with Persistent Illumination}

Existing \ac{sts} designs preserve traditional \ac{vbts} depth estimation by semi-transparent coatings~\cite{roberge2023stereotac,luu2023soft,luu2025vision}, switchable illumination~\cite{roberge2023stereotac}, or movable components~\cite{dong2025look}, but compromise visual clarity and require explicit, task-specific switching mechanisms. We adopt a fully transparent elastomer with persistent LED illumination, trading depth perception for continuous visual perception. The illumination includes a 24-bulb LED array (RGBW for \ac{vbts}-compatibility, or white-only for visual clarity) surrounding the elastomer. Diffusion films are applied to both the LEDs and the underlying acrylic plate surfaces to produce uniform, diffuse lighting and suppress interference from specular reflections.

Under this design, contact detection remains available through two mechanisms: light reflection changes at contact interfaces~\cite{hogan2022finger,athar2023vistac,zhang2023tirgel} and marker divergence from elastomer deformation~\cite{ablett2025multimodal}. This design explicitly prioritizes global contact state and visual context at the fingertips over fine-grained surface geometry reconstruction, aligning with recent literature identifying these modalities as critical for robust contact-rich manipulation~\cite{ablett2025multimodal,athar2023vistac,zhang2023tirgel,hogan2022finger,billard2019trends,heng2025vitacformer,yin2025learning,qi2023general}.

\subsection{Keyline Markers}

Elastomer transparency creates two marker detection challenges: (i) degraded detectability: conventional solid markers become invisible against matching backgrounds; (ii) noisy detections: environmental objects with blob-like appearances generate false detections. To address the detectability issue, we introduce \textit{keyline markers}: two concentric circles with contrasting colors, ensuring the inner edge remains visible as a detectable ``keyline'' regardless of background. 

As shown in \cref{fig:tacthru-umi-system}ab, the markers are fabricated by sequentially painting inner circles (black, $r_\text{in} = \SI{0.6}{mm}$) and outer circles (white, $r_\text{out} = \SI{1.0}{mm}$) using laser-cut masks. We deploy $N_m = 64$ markers with spacing $d_\text{marker}=$\SI{3.5}{mm} ($\tilde{d}_\text{marker}$ = \SI{51}{px} in pixel coordinate) on our \SI{40}{mm} $\times$ \SI{40}{mm} elastomer. Design constraints include: camera focus distance that enables both marker detection and visual perception, marker size that balances detectability with minimal visual occlusion, and spacing that exceeds the maximum marker deviation to prevent tracking ambiguity.

\subsection{Robust and Efficient Marker Tracking}
\label{sec:tracking_kf}

Although keyline markers solve the fundamental detectability problem, environmental noise and large contact deformations still cause tracking failures. We employ Kalman filtering~\cite{kalman1960new} for tracking each marker's state (position in image) $x_t \in \mathbb{R}^2$ at time $t$ based on its measurement (detected position) $z_t \in \mathbb{R}^2$ following:
\begin{equation}
    x_t = A_t x_{t-1} + w_t, \quad z_t = H_t x_t + v_t.
\end{equation}
\rev{We adopt a Random Walk model ($A_t = \mathbb{I}_2$) with direct position observation ($H_t = \mathbb{I}_2$). Process and measurement noises $w_t \sim \mathcal{N}(0, Q), v_t \sim \mathcal{N}(0, R)$ are parameterized by $Q = \sigma_w^2 \mathbb{I}_2, R = \sigma_v^2 \mathbb{I}_2$. Computationally, at timestep $t$, we perform the standard Kalman Filter steps: \textbf{(i) Predict} the prior state estimate $\tilde{x}_t$ and covariance $\tilde{P}_t$ (\cref{eq:kf:prd_x,eq:kf:prd_cov}). \textbf{(ii) Update} the posterior state $\hat{x}_t$ and covariance $\hat{P}_t$ based on measurement $z_t$ and computed Kalman Gain $K_t$ (\cref{eq:kf:gain,eq:kf:upd_x,eq:kf:upd_cov}). $\hat{x}_0, \hat{P}_0 = \epsilon \mathbb{I}_2$ are initialized based on sensor model with low uncertainty (\eg, $\epsilon=10^{-3}$).
\begin{align}
    \tilde{x}_t &= A_t \hat{x}_{t-1} &&= \hat{x}_{t-1} \label{eq:kf:prd_x} \\
    \tilde{P}_t &= A_t \hat{P}_{t-1} A_t^T + Q &&= \hat{P}_{t-1} + Q \label{eq:kf:prd_cov} \\
    K_t &= \tilde{P}_t H_t^T (H_t \tilde{P}_t H_t^T + R)^{-1} &&= \tilde{P}_t (\tilde{P}_t + R)^{-1} \label{eq:kf:gain} \\
    \hat{x}_t &= \tilde{x}_t + K_t (z_t - H_t \tilde{x}_t) &&= \tilde{x}_t + K_t (z_t - \tilde{x}_t) \label{eq:kf:upd_x} \\
    \hat{P}_t &= (\mathbb{I}_2 - K_t H_t) \tilde{P}_t &&= (\mathbb{I}_2 - K_t) \tilde{P}_t \label{eq:kf:upd_cov}
\end{align}
Measurement acquisition $z_t$ (marker's detected position) from sensor image $I_t$ involves:

\paragraph{Grayscale}: Conver the image to a grayscale iamge: $I_{\text{gs},t} \gets \texttt{Grayscale}(I)$.

\paragraph{Thresholding}: Set pixels to binary based on a threshold $\tau_i$ to reveal markers edges and suppress low-contrast background edges: $I_\text{bin,t}(u, v) \gets \mathbbm{1}(I_\text{gs,t}^{(u, v)} > \tau_i)$, where $(u, v)$ is pixel coordinate and $\mathbbm{1}(\cdot)$ is an indicator function .

\paragraph{Blob Detection}: Extract blob detections $Z_t : \{ z \} = \texttt{Blob}(I_{\text{bin,t}})$ with OpenCV \textit{SimpleBlobDetector}.

\paragraph{Data association:} Match the nearest blob for each marker as measurement: $z_t = \argmin_{z \in Z_t} \Vert z - \hat{x}_{t-1} \Vert$, which rejects potential environmental noises.

\rev{In extreme cases (\eg, catastrophic elastomer deformation and ambient lights), a few markers may become temporarily undetectable, and $z_t$ might be mis-matched to surrounding blobs. We enforce a marker motion continuity constraints: if the inter-frame displacement $\Vert z_t - \hat{x}_{t-1} \Vert > \tau_z$, we reject $z_t$ by setting $K_t = 0$ instead of~\cref{eq:kf:gain}, then the state estimate holds $\hat{x}_t = \hat{x}_{t-1}$ and the covariance accumulates ($\hat{P}_t = \hat{P}_{t-1} + Q \succeq \hat{P}_{t-1}$). $\tau_z$ should be selected to upper bound both the largest possible inter-frame marker displacement and half the marker spacing $\frac{\tilde{d}_\text{marker}}{2}$.}

To determine the parameters, we first calibrate $\sigma_v = 0.42$ px using sensor recordings with no marker motions, and empirically tune the noise ratio $\eta = \frac{\sigma_v}{\sigma_w} \approx 4$ (thus $\sigma_w = 0.11$ px) to balance tracking responsiveness with stability. $\tau_i = 0.784$ (200 in \texttt{UINT8}) and $\tau_z = $ 10px are determined empirically.}

\subsection{Evaluation on Marker Tracking}

\begin{figure}[!b]
    \centering
    \includegraphics[width=\linewidth]{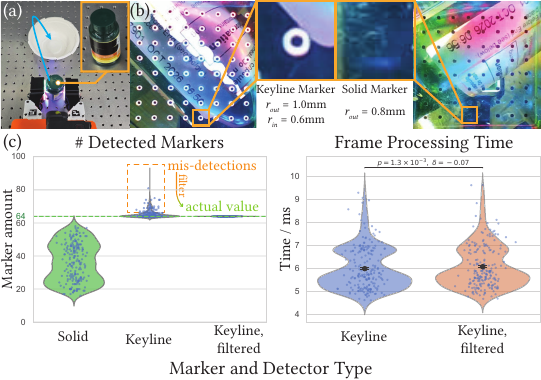}
    \caption{\textbf{Keyline marker design and filtering enable robust tracking.} (a) Evaluation setup compares two marker types (keyline \vs solid markers) during bottle grasping. (b) \GodDamnSensorName{} view comparison shows that keyline markers (left) remain distinct against complex backgrounds, while solid markers (right) become invisible. (c) Quantitative results demonstrate our filtered keyline method achieves stable tracking of all 64 markers while keeping efficiency (\SI{6.08}{ms} processing time), and the filtering step rejects false positives.}
    \label{fig:keyline_marker_design}
\end{figure}

We evaluate the proposed marker tracking algorithm by comparing keyline and solid marker designs. \cref{fig:keyline_marker_design}a shows two \GodDamnSensorName{} sensors mounted on the \ac{umi} data collector~\cite{chi2024universal}, one with keyline markers and one with solid black markers. \cref{fig:keyline_marker_design}b illustrates the challenge: solid markers become invisible against black backgrounds; keyline markers remain detectable but are subject to environmental noise.

We collect 8 trajectories (1628 frames per sensor), grasping a plastic bottle with complex black-and-white text. \GodDamnSensorName{} positions are swapped for half of the trajectories to ensure fair exposure to environmental conditions on both sides, and the bottle is rotated $90^\circ$ between collections, creating marker overlays with various areas around the bottle. We quantitatively compare three tracking approaches:
\begin{itemize}[leftmargin=*,nolistsep,noitemsep]
    \item \textbf{Solid:} Solid markers with standard blob detection.
    \item \textbf{Keyline:} Keyline markers with standard blob detection, matching detected blobs to nearest initial positions.
    \item \textbf{Keyline, filtered:} Keyline markers as above, with our Kalman filtering strategy (\cref{sec:tracking_kf}).
\end{itemize}

\cref{fig:keyline_marker_design}c presents quantitative results. Solid markers exhibit frequent missed detections, leading to incomplete tactile information. Keyline marker detection reports more than $N_m = 64$ markers due to environmental noises (orange dashed box). Our filtering strategy rejects these false positives, maintaining persistent tracking of all markers with an average latency of \SI{6.08}{ms} with low CPU overhead. This efficiency supports timely updates on the latest sensor observations, reducing temporal misalignment with other sensory modalities (\eg, vision and proprioception in~\cref{sec:tacthru_umi}) and saving time budget to enable high-frequency (\eg, \SI{120}{Hz}) operation.

\begin{figure}[!htb]
    \centering
    \includegraphics[width=\linewidth]{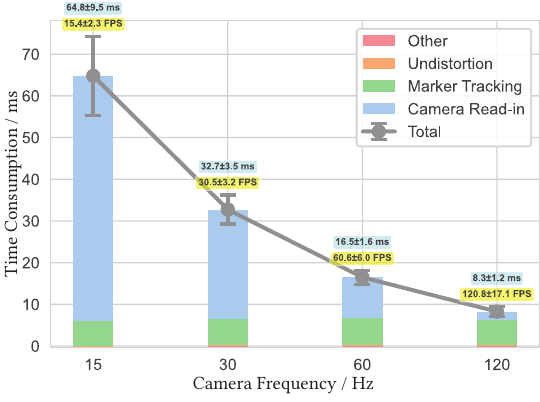}
    \caption{{Time consumptions of \GodDamnSensorName{}'s operation under various frequencies.} \GodDamnSensorName{} is tested in 15, 30, 60, and 120Hz, and the time consumption of each computational step is reported. Texts above error bars: iteration time consumption (blue) and equivalent FPS (yellow). The small overshoots in FPS are due to the actual on-board oscillator frequency and the UVC camera's buffering strategy.}
    \label{fig:proc_freq}
\end{figure}

We further measure the time consumption of each computation step during the sensor loop in~\cref{fig:proc_freq}, evaluated on a workstation (Ubuntu 22.04) with an AMD Ryzen 9 5950X CPU. The computational load of undistortion and marker tracking remains consistently low, and the process is mainly blocked at the image read-in step (\texttt{VideoCapture.read()} or \texttt{.retrieve()}) due to the UVC protocol and OpenCV behaviors. This confirms \GodDamnSensorName{}'s practical capability to operate at a high frequency.

\section{Learning Manipulation with \GodDamnSystemName{}}
\label{sec:tacthru_umi}

Although previous studies show invaluable applications of \ac{sts} sensors with task-specific controllers~\cite{yamaguchi2016combining,yamaguchi2017implementing,patel2018integrated,lancaster2019improved,hogan2022finger,lancaster2022optical,wang2022spectac,luu2023soft,athar2023vistac,roberge2023stereotac,luo2024compdvision,fan2024vitactip,dong2025look}, we apply \GodDamnSensorName{} to robotic manipulation based on imitation-learning, using its simultaneous tactile-visual perception for fine-grained and contact-rich tasks. We develop the \GodDamnSystemName{} imitation learning framework, extending the \ac{umi}~\cite{chi2024universal} and the Diffusion Policy~\cite{chi2025diffusion} with multimodal tactile-visual observations. Evaluations across multiple tasks highlight the sensor's multimodal perception capabilities, providing close-up visual observation of environments and objects along with tactile feedback, enabling more precise manipulation than traditional single-modality approaches.

\subsection{System Setup}

\paragraph*{Data collector}
The \GodDamnSystemName{} data collector adapts the \ac{umi}~\cite{chi2024universal} design, replacing standard fingers with \ac{sts} sensors on extended linkages (\cref{fig:tacthru-umi-system}c). \GodDamnSensorName{}s are connected to a computer via USB to stream real-time images during demonstrations and policy inference.

\paragraph*{Robot end-effector}
Although the \ac{umi} end-effector setup is compatible with multiple parallel grippers (\eg the Panda Hand), we design a low-cost gripper that directly mirrors the data collector's body and \GodDamnSensorName{} fingers (\cref{fig:tacthru-umi-system}c), with the finger width controlled by an Inspire LAS30-021D servo-electric cylinder ($\sim \$280$).

\subsection{Data Collection and Processing}

Our pipeline extends \ac{umi}~\cite{chi2024universal} with \GodDamnSensorName{} (\cref{fig:tacthru-umi-system}) and replaces the SLAM-based pose tracking with an HTC Vive Tracker for higher tracking success rates. While \GodDamnSensorName{} supports 120Hz operation with marker tracking, we use a sufficient and more widely-adopted~\cite{lambeta2020digit,liu2025vitamin} frequency of 30FPS. All data streams (wrist camera, tactile sensors, and proprioception) are synchronized to wrist-camera timestamps and stored in Zarr format for efficient training access.

\subsection{Policy Learning and Inference}

\begin{figure}[!b]
    \centering
    \includegraphics[width=\linewidth]{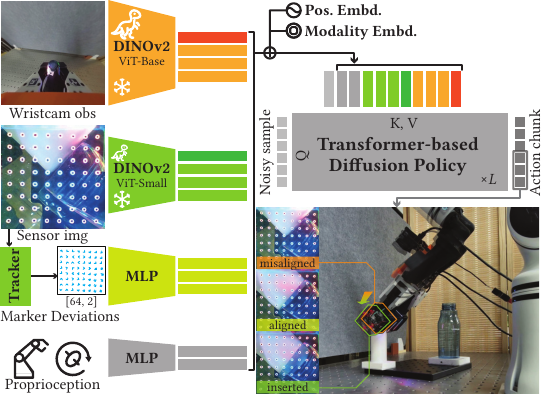}
    \caption{\textbf{Policy architecture.} Multimodal observations (wrist camera images, sensor images, marker deviations, and proprioception) are encoded as tokens, and used to condition a Transformer-based diffusion policy that denoises Gaussian noises into robot action chunks.}
    \label{fig:model_arch}
\end{figure}

We employ a Transformer-based Diffusion Policy~\cite{vaswani2017attention,chi2025diffusion} to learn mappings from multimodal observations to robot actions, allowing dynamic attention across simultaneously provided visual, tactile, and proprioceptive signals.

At timestep $t$, the observations include wrist-camera frames $\mathbf{I}_w^t :=\{I_w^i\}_{i=t-n^\text{obs}_w}^{t-1}$, sensor frames $\mathbf{I}_s^t := \{I_s^i\}_{i=t-n^\text{obs}_s}^{t-1}$, marker deviations $\Delta\mathbf{x}^t := \{\Delta x^{i, j}, j = 1, \dots, N_m\}_{i=t-n^\text{obs}_s}^{t-1}$, and proprioception $\mathbf{s}^t := \{ s^i \}_{i=t-n^\text{obs}_p}^{t-1}$ containing the pose of the end-effector and the width of the gripper in relative coordinates~\cite{chi2024universal}. Visual observations are encoded using DINOv2~\cite{oquab2023dinov2}: ViT-B for wrist cameras and ViT-S for \GodDamnSensorName{} frames ($14 \times 14$ patch size). Marker deviations and proprioception use dedicated \acp{mlp}. Each modality receives learnable embeddings ($z_w, z_s, z_x, z_p$) for transformer distinguishability:
\begin{align}
    \mathbf{z}_w & = \left\{ \text{DINO}_w(I) + z_w \vert I \in \mathbf{I}_w^t \right\}, \\
    \mathbf{z}_s & = \left\{ \text{DINO}_s(I) + z_s \vert I \in \mathbf{I}_w^t \right\}, \\
    \mathbf{z}_x & = \left\{ \text{MLP}_x(\Delta x) + z_x \vert \Delta x \in \Delta \mathbf{x}^t \right\}, \\
    \mathbf{z}_p & = \left\{ \text{MLP}_p(s) + z_p \vert s \in \mathbf{s}^t \right\}.
\end{align}

Concatenated tokens with positional embeddings condition the diffusion policy $\pi_\theta$, which denoises Gaussian samples into action chunks~\cite{zhao2023learning} $\mathbf{a} = \{ a^i \}_{i=t}^{t+T_a-1} \sim \pi_\theta (\mathbf{a} \vert \mathbf{z}_w, \mathbf{z}_s, \mathbf{z}_x, \mathbf{z}_p)$. Each action $a^i$ includes relative end-effector pose and gripper width targets. During execution, the first $L_a$ actions ($L_a \le T_a$) are sent to the robot controller for Cartesian space servoing.

\section{Experiments}

\begin{figure*}[t!]
    \centering
    \includegraphics[width=\linewidth]{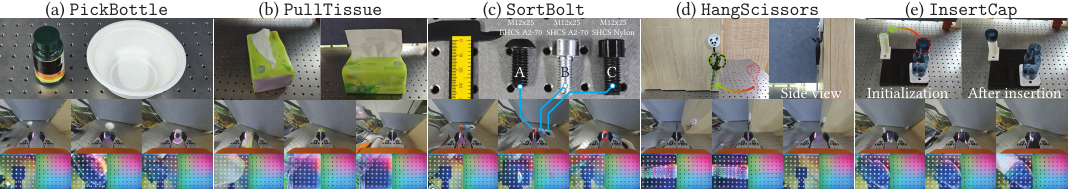}
    \caption{\textbf{Task demonstrations across five manipulation scenarios.} (a) \texttt{PickBottle}: basic pick-and-place, (b) \texttt{PullTissue}: thin-and-soft object manipulation, (c) \texttt{SortBolt}: visual discrimination, (d) \texttt{HangScissors}: tactile discrimination, (e) \texttt{InsertCap}: multimodal fusion. \textbf{Top:} Initial object configurations. \textbf{Middle:} Corresponding wrist-camera views. \textbf{Bottom:} Corresponding \GodDamnSensorName{} and GelSight images.}
    \label{fig:task_settings}
\end{figure*}

\begin{figure}[!b]
    \centering
    \includegraphics[width=\linewidth]{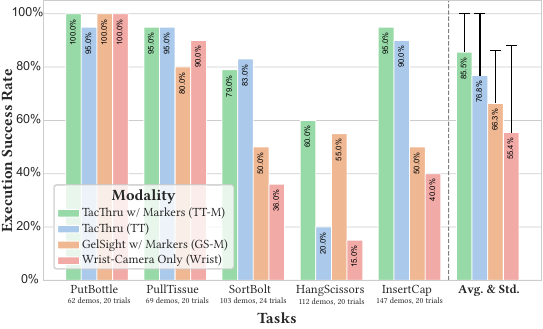}
    \caption{\textbf{Quantitative results across manipulation tasks and sensing modalities.} \texttt{\textbf{TT-M}}: \GodDamnSensorName{} with markers, \texttt{\textbf{TT}}: \GodDamnSensorName{} images, \texttt{\textbf{GS-M}}: GelSight with markers, and \texttt{\textbf{Wrist}}: vision-only.}
    \label{tab:quant_res}
\end{figure}

\begin{figure*}[t!]
    \centering
    \includegraphics[width=\linewidth]{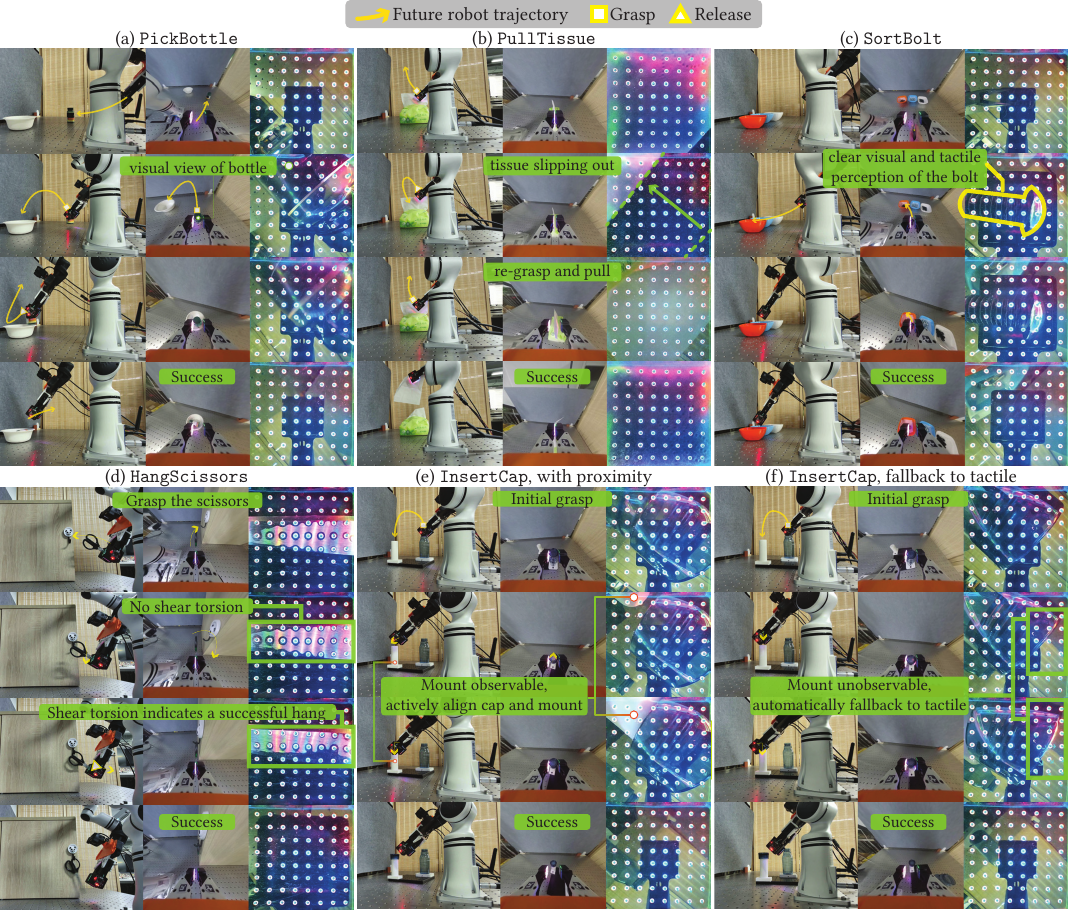}
    \caption{\textbf{Representative policy rollouts.} Each column (a-f) shows the temporal progression of a task execution (top down), with synchronized views: third-person view (left), wrist camera (center), and \GodDamnSensorName{} with marker deviations overlaid (right; 4× magnified for visibility). Annotations highlight key manipulation phases and sensing feedback. See supplementary video for additional rollouts across all settings.}
    \label{fig:exp_res}
\end{figure*}

We evaluated \GodDamnSystemName{} across five manipulation tasks spanning pick-and-place, sorting, and insertion scenarios. These tasks systematically assess different sensing modalities: tactile information (contact events and shear forces), visual perception (object and environment observation), and their simultaneous operation. Our key finding is that \GodDamnSensorName{}'s multimodal feedback enriches perception of object appearance, state, and the environment throughout the entire manipulation process, even before contact or when handling extremely thin and soft objects. This allows policies to leverage detailed environmental cues for enhanced fine-grained and contact-rich manipulation while maintaining inference efficiency.

\subsection{Task Settings}

\cref{fig:task_settings} illustrates the experimental setups. Each column shows a task with the object to manipulate (first row), wrist-camera view during demonstration (second row), and the corresponding \GodDamnSensorName{} and \ac{vbts} images (third row).

\paragraph*{\texttt{PickBottle}}
A bottle and a bowl are placed randomly. The robot must grasp the bottle and place it in the bowl. This basic pick-and-place task validates \GodDamnSystemName{}'s effectiveness of imitation learning and real-world inference.

\paragraph*{\texttt{PullTissue}}
A tissue pack is placed randomly, and the robot must grasp and fully extract a single tissue. Standard tactile sensors struggle to detect contact with thin, soft paper, making the visual modality of \ac{sts} crucial for this task.

\paragraph*{\texttt{SortBolt}}
One of three M12×25 bolts is placed between fingertips. The robot must grasp the bolt and place it in the corresponding bowl. Bolts vary in head shape (A: button head; B, C: socket head) and color (A, C: black; B: silver). The global wrist-camera view cannot distinguish these small bolts, while traditional tactile sensing cannot separate geometrically identical but differently colored bolts. \GodDamnSensorName{} allows for distinguishing both color and shape through \ac{sts} perception.

\paragraph*{\texttt{HangScissors}}
The scissors are placed between the fingertips; the robot must grasp and hang them on a hook~\cite{liu2025vitamin}. This requires tactile feedback to distinguish successful hanging (triggers gripper release) from missed attempts (triggers retry).

\paragraph*{\texttt{InsertCap}}
The robot must grasp a bottle cap and insert it onto a mount, which requires millimeter-level precision alignment. Although insertion tasks are typically based on tactile signals~\cite{xu2023efficient,chen2024general}, \GodDamnSensorName{} enables direct visual servoing for the cap-mount alignment.

\subsection{Experimental Setup}

\paragraph*{Setup and policy variants}
For fair comparison, we equip the gripper with \GodDamnSensorName{} on one finger and a GelSight-type sensor on the other, ensuring identical training trajectories across modality variants. We collect 62--147 demonstrations per task (\cref{tab:quant_res}) and train four policy variants that form ablations and comparisons. Of note, all policies include a wrist camera ($\mathbf{I}_w^t$) and proprioception ($\mathbf{s}^t$) as the base input.
\begin{itemize}[leftmargin=*,nolistsep,noitemsep]
    \item{\textbf{\texttt{TT-M}:}} \GodDamnSensorName{} images ($\mathbf{I}_s^t$) and marker deviations ($\Delta \mathbf{x}^t$).
    \item{\textbf{\texttt{TT}:}} \GodDamnSensorName{} images only ($\mathbf{I}_s^t$); markers are visible but not explicitly tracked and provided (as an ablation).
    \item{\textbf{\texttt{GS-M}:}} GelSight images (rectified by the idle image for isolating contact) and marker deviations (tactile baseline).
    \item{\textbf{\texttt{Wrist}:}} Wrist camera only ($\mathbf{I}_w^t$) (vision-only baseline).
\end{itemize}

\paragraph*{Training and evaluation}
We use observation horizons $n^\text{obs}_w = 1, n^\text{obs}_s = 1, n^\text{obs}_p = 2$, predict $T_a = 16$ action steps, and execute the steps 3-8. Each policy is trained for 150 epochs using the AdamW optimizer~\cite{loshchilov2019decoupled} (learning rate = $3 \times 10^{-4}$, one-cycle scheduler). We evaluate each checkpoint with 20 rollouts (24 for \texttt{\textbf{SortBolt}}; 8 for each bolt) with randomized initialization that matched the data-collection distribution.

\subsection{Results}

\paragraph*{Pick-and-place}
\rev{In \textbf{\texttt{PickBottle}} (\cref{fig:exp_res}a), all policies achieve near-perfect success ($\geq 95\%$), validating that \GodDamnSystemName{} effectively integrates multimodal signals without compromising basic manipulation. This establishes a baseline for specialized sensing in more challenging tasks.}

\paragraph*{Thin-and-soft object perception}
\textbf{\texttt{PullTissue}} reveals the limitations of conventional tactile sensing: thin-and-soft exert minor pressure that is hardly detected. \GodDamnSensorName{} overcomes this through direct visual observation of the inter-finger workspace, providing continuous feedback on tissue position. \rev{When tissue slippage (\cref{fig:exp_res}b) occurs due to insufficient grip force, while \ac{vbts} (\texttt{\textbf{GS-M}}) and vision-only (\texttt{\textbf{Wrist}}) policies fail to notice this accident, \GodDamnSensorName{} immediately detects the displacement and triggers corrective retry.}

\begin{figure}[t!]
    \centering
    \includegraphics[width=\linewidth]{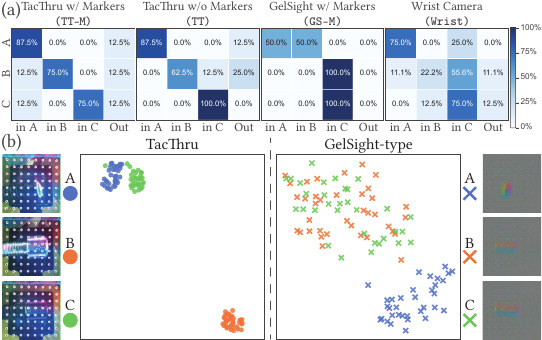}
    \caption{\textbf{Bolt sorting performance analysis.} (a) Confusion matrices of placement accuracies. Rows: ground-truth bolt type; columns: executed placement (``Out'': dropping outside). \GodDamnSensorName{} policies (\texttt{\textbf{TT-M}}, \texttt{\textbf{TT}}) achieve high performance, while \texttt{\textbf{GS-M}} confuses geometrically identical bolts B and C. (b) t-SNE visualization of sensor images' DINOv2 CLS token. \GodDamnSensorName{} produces clearly separated clusters, while GelSight embeddings for bolts B and C overlap.}
    \label{fig:sort_confusion}
\end{figure}

\paragraph*{Visual discrimination}
\textbf{\texttt{SortBolt}} (\cref{fig:exp_res}c) demonstrates \GodDamnSensorName{}'s fine-grained visual discrimination of the object. The small M12×25 bolts present significant challenges: wrist cameras cannot resolve geometric details at mounted distances, while identical bolt shapes make purely tactile discrimination impossible. \GodDamnSensorName{}'s close-proximity view captures fine geometric features and subtle color differences. \rev{\cref{fig:sort_confusion}b provides quantitative evidence through t-SNE visualization of DINOv2 \texttt{\textbf{CLS}} embeddings: \GodDamnSensorName{} tokens are clearly separated, while \ac{vbts} embeddings are more overlapped, which are aligned with \GodDamnSensorName{}'s $\approx 80\%$ accuracy (\texttt{\textbf{TT-M}}, \texttt{\textbf{TT}}) and \texttt{\textbf{GS-M}}'s obfuscation of bolts B and C (\cref{fig:sort_confusion}a).}

\paragraph*{Tactile discrimination}
\textbf{\texttt{HangScissors}} (\cref{fig:exp_res}d) exemplifies scenarios where visual observation alone cannot determine task completion: the wrist camera cannot reliably identify scissors-handle engagement due to 2D perception and occlusion. \rev{Marker displacements (\texttt{\textbf{TT-M}}, \texttt{\textbf{GS-M}}) provide tactile evidence of successful hang, vital to the precise timing of the gripper release, preventing premature release or continued downward forces. This justifies the rationale of explicitly extracting the marker deviations of tactile signals.}

\paragraph*{Multimodal fusion}
\textbf{\texttt{InsertCap}} (\cref{fig:exp_res}e-f) presents the unique advantage of \GodDamnSensorName{}: simultaneous access to visual and tactile information enables the selection of adaptive strategies. When the cap-mount interface remains visible (15 / 20), the policy employs visual servoing, directly aligning visual features for precise insertion (\cref{fig:exp_res}e). However, when the grasp degrades this visual perception (5 / 20), the policy seamlessly returns to tactile-based insertion, using marker displacement patterns to detect contact and guide alignment (\cref{fig:exp_res}f). This adaptive behavior is enabled by (i) the \GodDamnSensorName{}'s \textit{simultaneous} visual-tactile perception, and (ii) Transformer-based policy's capability to properly attend to multimodal signals, acquired from the demonstration data without explicit rule-based programming.

\subsection{Discussions}

Our experimental results reveal three key insights on tactile-visual perception for learning multimodal robot manipulation.

\begin{figure}
    \centering
    \includegraphics[width=\linewidth]{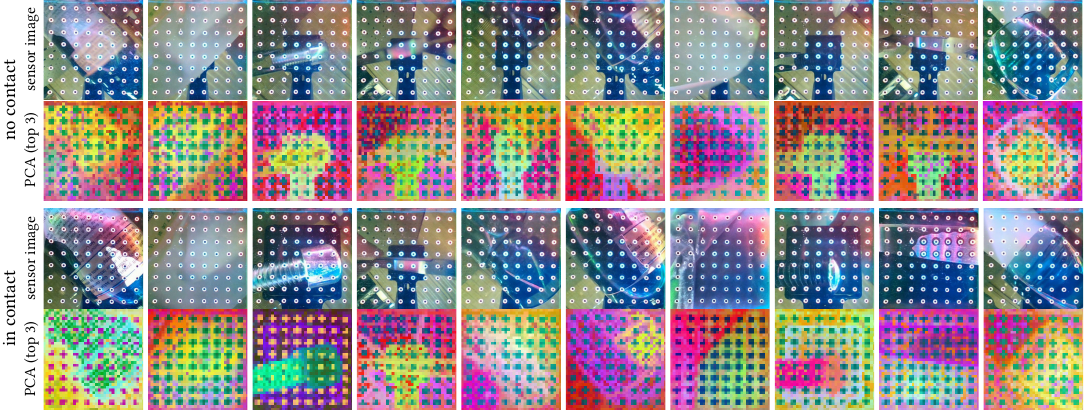}
    \caption{\textbf{\ac{pca} heatmaps of DINOv2 image tokens.} Colors are computed by normalizing patch tokens' coefficients of the first three principal components to RGB space.}
    \label{fig:dino_v2}
\end{figure}

\rev{\paragraph*{Encoding \GodDamnSensorName{} images with DINOv2} Despite the domain gap introduced by markers and the elastomer, we find DINOv2 effective in capturing meaningful features from \GodDamnSensorName{} images for downstream policies. This performance stems from DINOv2's strong generalization capability, thanks to its discriminative self-supervision training on a large, diverse, and curated dataset (LVD-142M)~\cite{oquab2023dinov2}, facilitating zero-shot transfer to unseen image domains~\cite{oquab2023dinov2,song2025dino,caraffa2024freeze,waldmann2025panopticon} in practice. We further validate this effectiveness by performing \ac{pca} on DINOv2 patch tokens extracted from our demonstration data (\cref{fig:dino_v2}), covering both contact and non-contact scenarios. By mapping the first three principal components' coefficients to RGB channels~\cite{oquab2023dinov2}, we visualize the token similarities of the patch tokens. The resulting heatmaps demonstrate that DINOv2 distinctively separates markers, manipulated objects, and background elements (\eg, the opposing finger), despite minor noises. This confirms that DINOv2 is capable of providing clear, structured cues for downstream policy learning.}

\paragraph*{Adaptive multimodal strategies}
Policies trained with \GodDamnSensorName{} naturally learn to weight sensing modalities based on the current context, as demonstrated in \texttt{\textbf{InsertCap}}, where the same policy employs vision-based alignment when visible and tactile-based insertion when occluded. This adaptive behavior emerges without explicit programming, suggesting fundamental advantages of simultaneous over sequential sensing.

\paragraph*{Overcoming conventional tactile limitations}
\GodDamnSensorName{} uniquely handles scenarios where traditional tactile sensors fail. \Eg, thin objects such as tissue generate insufficient forces for tactile sensors but remain visually observable through \GodDamnSensorName{}. This expands the range of manipulable objects for tactile robotics.

\paragraph*{Practical deployment viability}
Despite significant domain differences due to transparent elastomer, marker overlays, and contact deformations, standard pre-trained visual encoders prove sufficient for robust policy learning. This finding substantially reduces implementation barriers and suggests that \GodDamnSensorName{} can be integrated into existing vision-based manipulation pipelines with minimal modification.

\paragraph*{Sensor Failure Modes} While \GodDamnSensorName{} robustly handles nominal contact forces, it shares specific failure modes with other elastomer-based sensors, being susceptible to elastomer delamination under extreme loads or sharp indentation. Integrating reinforced surface materials (\eg, stretchable protective layers~\cite{zhao2025polytouch}) represents a viable solution for enhancing durability. Besides, extremely high-intensity ambient illumination can interfere with the camera's auto-exposure, reducing marker contrast and causing missed detection. While it can be mitigated by manually decreasing the thresholding parameter $\tau$, we identify the adaptive exposure control as a valuable feature for future research.

\section{Conclusions}

We introduce \GodDamnSensorName{}, an \ac{sts} tactile sensor that provides simultaneous tactile and visual perception through transparent elastomer, persistent illumination, and keyline marker tracking. Integrated within our \GodDamnSystemName{} imitation learning platform, it demonstrates superior performance across various manipulation tasks, addressing the fundamental limitations of existing tactile sensing while maintaining compatibility with standard vision pipelines.

The accessible design and open-source \GodDamnSystemName{} platform position this work as a practical enhancement for the manipulation research community. Future directions include large-scale data collection combined with synthetic tactile simulation~\cite{li2025taccel,chen2024general,xu2023efficient} to support pre-training of specialized encoders, and exploration of complex dexterous tasks that fully leverage \GodDamnSensorName{}'s simultaneous sensing capabilities.

{
\bibliographystyle{ieeetr}
\balance
\bibliography{reference_header,reference}
}

\end{document}